\documentclass{article}
\usepackage{spconf,amsmath,graphicx}

\usepackage{algorithm}
\usepackage{algpseudocode}
\usepackage{url} 
\usepackage{booktabs} 
\usepackage{xcolor} 
\usepackage{todonotes} 
\usepackage{multirow} 

\usepackage{color, colortbl}
\newcommand{\norm}[1]{\left\lVert#1\right\rVert}

\title{Improving Membership Inference in ASR Model Auditing \\with Perturbed Loss Features}
\name{Francisco Teixeira~$^{*1}$, Karla Pizzi~$^{*2,3}$, Raphael Olivier~$^{*4}$, Alberto Abad~$^1$, Bhiksha Raj~$^4$, Isabel Trancoso~$^1$
\thanks{$^{*}$Authors contributed equally to this work, and are sorted alphabetically.\\This work was funded in part by Portuguese national funds through Fundação para a Ciência e a Tecnologia, with reference DOI:10.54499/UIDB/50021/2020; by the Recovery and Resilience Plan and Next Generation EU European Funds, with reference C644865762-00000008 Accelerat.AI; the Bavarian Ministry of Economic Affairs, Regional Development, and Energy.}
}
\address{$^1$INESC-ID/Técnico, University of Lisbon, Portugal, $^2$Technical University Munich, Germany,\\$^3$Fraunhofer AISEC, Germany, 
$^4$LTI, Carnegie Mellon University, USA}
\begin{document}
\ninept
\maketitle
\begin{abstract} 
\vspace{-0.2cm}
Membership Inference (MI) poses a substantial privacy threat to the training data of Automatic Speech Recognition (ASR) systems, while also offering an opportunity to audit these models with regard to user data.
This paper explores the effectiveness of loss-based features in combination with Gaussian and adversarial perturbations to perform MI in ASR models.
To the best of our knowledge, this approach has not yet been investigated. 
We compare our proposed features with commonly used error-based features and find that the proposed features greatly enhance performance for sample-level MI. 
For speaker-level MI, these features improve results, though by a smaller margin, as error-based features already obtained a high performance for this task. 
Our findings emphasise the importance of considering different feature sets and levels of access to target models for effective MI in ASR systems, providing valuable insights for auditing such models.

\end{abstract}

\section{Introduction}
\label{sec:intro}
\vspace{-0.1cm}
Automatic Speech Recognition (ASR) systems are revolutionising the way we interact with technology. 
The recent progress of ASR systems has led to the deployment of numerous cloud-based services and applications that leverage speech as a means of human-computer interaction. 
An estimated 4.2 billion voice assistants were in use worldwide in 2020~\cite{statista2022number}, and the smart-speaker global market share is expected to reach 35.5 billion US dollars by~2025~\cite{statista2023smart}.

At the same time, the use of these systems has given rise to concerns regarding user privacy, primarily due to the privacy vulnerabilities of machine learning (ML) models~\cite{feng2023review}. 
Of particular concern are Membership Inference (MI) attacks, which exploit the susceptibility of ML models to 
reveal whether individuals were included in the model's training dataset, thereby disclosing potentially sensitive information~\cite{shokri2017membership}.
For instance, if one knows specific characteristics of the population that make up the training set -- e.g., a training dataset that consists only of individuals affected by a certain illness -- it follows that an individual that is part of this dataset will share these potentially sensitive characteristics~\cite{shokri2017membership}.

Even though MI is most often considered an attack on the privacy of learning data, it may equally be seen as a tool to protect data donors and service providers.
As an auditing tool, MI can provide evidence that available models do not leak information about their training data and show that the audited model is in adherence to data protection regulations~\cite{ye2022enhanced}, such as the European Union's General Data Protection Regulation (GDPR)~\cite{gdpr}, or California's Consumer Protection Act (CCPA)~\cite{ccpa}. 
MI can also be used to audit service providers' use of customer data. 
Specifically, a service provider that trains an ML model with user data without adequate consent may violate data protection regulations. 
In this case, MI can be used 
to assert whether or not a data sample was used during training, protecting both users and service providers~\cite{song2019auditing, miao2021auditor,li2023membership}.

MI is, thus, an important aspect of trustworthy machine learning, that should be studied in all its facets and for all types of data.
However, while MI has been extensively studied in the realms of image and text data~\cite{hu2022membership}, the focus on speech data, particularly in what concerns ASR models, remains limited~\cite{chen2023slmia,kong2023efficient,miao2021auditor, shah2021evaluating, tseng2022membership, li2023membership}.

Most of the scarce literature on MI in ASR has focused on the use of transcription errors~\cite{shah2021evaluating}, transcription-reference similarity scores~\cite{miao2021auditor}, or both~\cite{li2023membership}, as features to classify membership. 
Tseng et al.~\cite{tseng2022membership} is an exception to this and explore MI in self-supervised speech models, using frame-similarity scores instead.
MI in ASR is also considered under different target use cases: Shah et al.~\cite{shah2021evaluating} and Tseng et al.~\cite{tseng2022membership} view their work as a traditional MI attack, targeted at understanding the vulnerabilities of ASR models,
whereas Miao et al.~\cite{miao2021auditor} and Li et al.~\cite{li2023membership} pose their work from an auditing perspective, where MI is a tool to check for unauthorised use of~data.

All of these works, however, have a strict adherence to \textit{black-box} scenarios, which, in the case of \cite{miao2021auditor,shah2021evaluating,li2023membership} means that only processed (i.e., decoded) model outputs are available, and consequently, only error-based features are used. 
Contrarily, we argue that having access to the model's output logits is a reasonable assumption that should be explored.
We consider this to be particularly true in auditing scenarios, where service providers are under scrutiny for potentially having trained their model on user data without consent, and are required to provide some level of model access to the auditor.

In this study, we focus on the auditing scenario for ASR models. 
We consider grey- to white-box access to the model, specifically: access to the raw output of the ASR model and some knowledge of the training data distribution (grey-box), as well as the ability to back-propagate through the model (white-box).
Our focus also extends beyond sample-level MI to include speaker-level MI, i.e., inferring whether an individual's data was part of the model's training data, without knowing the exact samples that were used for this~purpose.

Under these assumptions, we explore loss information (i.e., Kullback-Leibler (KL) divergence and Connectionist Temporal Classification (CTC) loss) when performing MI, which, to the best of our knowledge, no previous work on the topic of MI for ASR has used. 
To gain more information about the decision boundary surrounding a given utterance, we further enrich these features by computing the losses over two types of input perturbations: Gaussian noise and adversarial noise. 
Similar perturbations have been explored in other domains, though using different protocols~\cite{choquette2021label, Rezaei2021OnTD}.

We conduct our experiments with Transformer~\cite{vaswani2017attention} and Conformer models~\cite{gulati2020conformer}, trained on subsets of LibriSpeech~\cite{librispeech}.
We observe that loss features outperform error features at sample-level MI, particularly when combined with the proposed perturbations. 
At speaker-level MI, we observe closer results for both sets of features with loss features still being able to achieve higher performances\footnote{The results of Table~\ref{tab:transLS360_sam_features} and \ref{tab:aggregated_results_sam} have been updated in comparison to the original paper at the ICASSP 2024 workshop on Trustworthy Speech Processing. 
We kindly ask researchers to refer to the updated numbers.}.

The remainder of this document is organised as follows: in Section~\ref{sec:features}, we describe our methodology and proposed features and perturbations; in Section~\ref{sec:expe}, we describe the experimental setup; and in Section~\ref{sec:results} we present and discuss the results obtained. 
Finally, Section~\ref{sec:conclusions} presents our conclusions and topics for future work.

\vspace{-0.2cm}
\section{Methodology}
\vspace{-0.25cm}
\label{sec:features}


To perform membership inference, we apply a methodology similar to previous works~\cite{shokri2017membership,shah2021evaluating,li2023membership}. 
Given a \textit{target} model to perform MI, we first train a \textit{shadow} model on a dataset that is disjoint from that of the target model.
We then build a balanced binary classification dataset of input utterances labelled positively \textit{iff} they are in the shadow model's training set.
The set of speakers for the positive and negative samples is the same, to ensure our classifier is distinguishing between seen and unseen samples, and not between seen and unseen speakers. 
We train a binary classifier for MI on this dataset, using the features 
described in the remainder of this section. 
As a final step, this binary classifier is used to evaluate a test set of utterances
with regard to their membership in the training set of the target model. 
We refer to this process as \textit{sample-level} MI. 
To perform \textit{speaker-level}, utterances are labelled positively \textit{iff} their \textit{speaker} was in the training set.
To ensure that the MI classifier is recognising speaker membership and not sample membership, we make sure that positive samples are not part of the ASR model's training data.
In what follows, we present the three feature categories that were used in our MI framework.

\vspace{-0.2cm}
\subsection{Baseline: error features}
\vspace{-0.2cm}
Our baseline feature extractor corresponds to a set of errors computed between the target and output transcriptions of the ASR model, combined with the model's confidence for these transcriptions.
Specifically, we use the word error rate (WER); the length-normalised counts for edits, substitutions, insertions and deletions; the length ratio between prediction and target transcription; and the confidence of the model regarding the transcription.
This is inspired by the best-performing set of the features evaluated in~\cite{shah2021evaluating}.
We compute all these features for the top-4 transcription hypotheses of the model and dub their combination as the \emph{errors} feature set. 

\vspace{-0.2cm}
\subsection{Loss-based features}
\vspace{-0.2cm}
The main focus of this work is the set of features that can be computed from the non-processed (i.e., non-decoded) output logits of the model.
We consider that these features contain a higher amount of information on membership than features computed from a post-processed output, as long as they are properly modelled~\cite{carlini2021membership}.
As loss-based features, we consider the losses used to train a transformer-based ASR model: the attention loss, which corresponds to the KL divergence between the output log-probabilities and the target transcription; and the CTC loss~\cite{graves2006connectionist}.
\vspace{-0.2cm}

\subsection{Perturbed features}
\vspace{-0.2cm}
To characterise the decision boundary around a given data point and to potentially improve the MI decision, we extend the loss features by perturbing the input signal using Gaussian and adversarial noise.

\vspace{-0.25cm}
\paragraph*{Gaussian noise}
Inspired by~\cite{jayaraman2020revisiting, choquette2021label}, we perturb input data with random Gaussian noise. 
Gaussian perturbations are agnostic to the model and data, and let us evaluate the model's ``average'' behaviour when getting further away from the input in arbitrary directions.
We use decreasing levels of the signal-to-noise ratio (SNR), moving the perturbed signal away from the original input. 
This set of perturbed signals is then fed to the ASR model, from whose output we compute our set of MI features.
Since using a single perturbation per SNR value would only give us information on the decision boundary regarding one random direction, for each SNR value, we select multiple random perturbations. 
The MI features computed from these random perturbations for the same value of SNR are then summarised by their mean and standard deviation.
This procedure is summarised in Algorithm~\ref{alg:gauss}.
\vspace{-0.2cm}

\begin{algorithm}[h]
\footnotesize
\begin{algorithmic}[1]
\Require Input $x$, set of SNRs $S$, $\#$runs $N$, model $M(\cdot)$, target transcription $y$, feature extractor $F(\cdot)$
\State $\text{feats} \gets [\,]$
\For{ $\text{snr} \in S$ }
    \State $\text{feats}_\text{snr} \gets [\,]$
    \For{$n \leq N$}
     \State $\delta \sim \mathcal{N}(0, I)$ \Comment{Sample Gaussian noise}
     \State $\delta_{\text{snr}} \gets \sqrt{\frac{\norm{x}_2^2}{\text{snr}\times\norm{\delta}_2^2}}\times\delta$ \Comment{Scale  noise to SNR}
     \State $ \text{feats}_\text{snr}[n] \gets F(M(x + \delta_{\text{snr}}), y)$
    \EndFor
    \State $\text{feats}[\text{snr}] \gets (\text{mean}(\text{feats}_\text{snr}), \text{stddev}(\text{feats}_\text{snr}))$
\EndFor
\State  \Return \text{feats}
\end{algorithmic}
\caption{Gaussian noise-based feature computation.}
\label{alg:gauss}
\end{algorithm}
\vspace{-0.3cm}
\paragraph*{Adversarial noise}
In addition to random perturbations, we propose to explore "worst-case" directions, for which the decision boundary is near the data point. 
Contrary to~\cite{Rezaei2021OnTD, zhang2022evaluating}, we do not estimate the ``distance to the decision boundary'' (an ambiguous notion for transduction tasks), but rather find directions of maximal error given a fixed perturbation budget. 
To do this, we run a panel of adversarial attacks, i.e., algorithms that find small perturbations of inputs that can fool ML models into changing their decisions. Details about adversarial attacks can be found in~\cite{goodfellow2016explaining}, and in~\cite{abdullah2020sok} for ASR in particular. 
We focus on the untargeted Projected Gradient Descent (PGD) attack~\cite{madry18} in the $L_\infty$ norm, a standard for adversarial perturbations. 
Given a radius $\epsilon$, we compute $N$ gradient steps of step size $\eta$, and at every step clip the perturbation so that $\norm{\delta}_\infty\leq\epsilon$. We apply this attack with different radii, and compute features for all returned perturbations. 
We detail this procedure in Algorithm \ref{alg:adv}.
Since it is necessary to perform back-propagation through the model to create the adversarial perturbations, the use of these features entails white-box model access.

\begin{algorithm}[ht]
\footnotesize
\begin{algorithmic}[1]
\Require Input $x$, set of radii $\mathcal{E}$, number of steps $N$, step size $\eta$, model $M(\cdot)$, target transcription $y$, feature extractor $F(\cdot)$
\State $\text{feats} \gets [\,]$
\For{ $\epsilon \in \mathcal{E}$ }
    \State $\delta_{\epsilon} \sim \mathcal{U}(-\epsilon I,\epsilon I)$
    \For{$n \leq N$}
     \State $g= \text{sign}\left(\frac{d}{d\delta_{\epsilon}}L(M(x+\delta_{\epsilon}),y)\right)$ \Comment{Gradient}
     \State $\delta_{\epsilon} \gets \text{clip}_\epsilon(\delta_{\epsilon}+\eta g)$ \Comment{Optimization \& projection}
    \EndFor
     \State $\text{feats}[\epsilon] \gets F(M(x + \delta_{\epsilon}), y)$
\EndFor
\State \Return $\text{feats}$
\end{algorithmic}
\caption{Adversarial-based feature computation.}
\label{alg:adv}
\end{algorithm}

\section{Experimental setting}
\label{sec:expe}
\subsection{Experiments}
\label{sec:models}
\vspace{-0.1cm}
For the experiments in this work, we trained three ASR models, varying in training data and architecture:
\begin{enumerate}
    \item An encoder-decoder transformer model (\textbf{T1}) \cite{speechbrain};
    \item A transformer model trained on a disjoint set of data but from the same distribution as the data used to train T1 (\textbf{T2});
    \item A conformer model~\cite{gulati2020conformer} trained on the same data as T2 (\textbf{C1}).
\end{enumerate}

To validate our hypothesis -- \textit{loss-based features improve upon error-based features} -- we performed a comparative and ablative study over the feature sets described in the previous section. 
Specifically, we compared the performance of the \textit{errors} feature set with the \textit{loss} feature set and with the combination of the losses with the Gaussian perturbations, adversarial perturbations, and both. 
To have an upper bound on the performance of each feature set, in this ablation study, the shadow and target models were the same.
To this end, we used \textbf{T1} as both the shadow and target model -- \textbf{T1} to \textbf{T1} (where ``A to B'' denotes A as the shadow model and B as the target model).

In addition to the above, when performing MI, it is reasonable to consider that different model architectures and models trained on different datasets will behave differently regarding the training losses and output errors.
As such, we performed two additional experiments: \textbf{T2} to \textbf{T1}; and \textbf{C1} to \textbf{T1}.
Experiment \textbf{T2} to \textbf{T1} corresponds to the case where the model's architecture is known, while experiment \textbf{C1} to \textbf{T1} corresponds to the case where the model's architecture is not known. 
In both cases, the training data of the shadow models (\textbf{T2} and \textbf{C1}) is different from that of the target model \textbf{T1}, but comes from the same data distribution.
These two experiments emulate auditing settings where access to and knowledge of the target model is limited, providing information about the behaviour of the proposed features in these harder but more realistic scenarios.

\vspace{-0.1cm}
\subsection{Data}
\vspace{-0.1cm}
The datasets used to train the ASR target and shadow models, as well as to train the MI classifiers, are built from data taken from LibriSpeech (LS)~\cite{librispeech}. 
More specifically, the dataset used to train \textbf{T1}, our target ASR model, was composed of 300h from LS's \textit{train-clean-360} partition. 
Similarly, the dataset used to train models \textbf{T2} and \textbf{C1} was composed of 80h from LS's \textit{train-clean-100} partition. 
The datasets used to train the MI classifier were composed of 5,000 utterances, whereas the test set contains 1,000 utterances. 
All datasets were balanced in terms of positive and negative samples.

\vspace{-0.1cm}
\subsection{Attack perturbations}
\vspace{-0.1cm}
We experimented with several choices of hyper-parameters for the perturbations detailed in Algorithms \ref{alg:gauss} and \ref{alg:adv}. 
The parameters reported in this section correspond to those that performed the best for held-out data.
For Gaussian perturbations, we use eight different SNRs linearly spaced between 0 and $50\text{dB}$. 
For each SNR, we perturb the signal four times, after which we average the resulting features.
For adversarial perturbations, we use 16 adversarial radii $\epsilon$: nine evenly spaced from 0.001 to 0.009 and seven from 0.01 to 0.07. 
We fix $\eta=1$ and $N=1$, which we find are as effective for MI as more computationally expensive hyper-parameters.

\subsection{Evaluation metrics}
\vspace{-0.1cm}
To allow our work to be compared to other approaches in the literature, our main metrics of evaluation are Accuracy (Acc) and Area Under the ROC Curve (AUC).
However, as argued by~\cite{carlini2021membership}, in MI in general, and in auditing scenarios in particular, the ``cost'' of deciding that a sample is in the training set -- while it should not be in it -- is much higher than deciding that the sample is not in the training set. 
In line with this argument, we also report the performance of our classifiers in terms of the True Positive Rate (TPR) obtained for two very low false positive rates (FPR): $0.1$ and $0.01$.

\vspace{-0.2cm}
\subsection{Implementation details}
\vspace{-0.15cm}
All ASR models were trained using SpeechBrain~\cite{speechbrain} and followed the default configurations and training parameters, except for: the number of epochs = 60; batch size = 16; gradient accumulation factor = 2. 
All experiments were performed without the use of a language model. 
\textbf{T1} obtains a WER of 5.45\% for LS's test-clean and 15.17\% for LS's test-other partitions; \textbf{T2} obtains WERs of 10.32\% and 24.70\%; and \textbf{C1} obtains WERs of 6.23\% and 16.77\%. 
When computing error features, decoding was performed with a beam size of 30. 
Experiments using adversarial noise were built with the \textit{robust-speech} package~\cite{Olivier2022RI}.
MI is performed using Scikit-Learn's Random Forest (RF) classifier~\cite{scikit-learn} with 100 estimators.
RF scores correspond to the mean of the predicted class probabilities for all decision trees in the RF; predictions are made with a 0.5 threshold.

\vspace{-0.1cm}
\section{Results}
\label{sec:results}

%


\vspace{-0.1cm}
\begin{table*}[ht]
\setlength\tabcolsep{3 pt}
    \centering
    \begin{tabular}{l|rrrr|rrrr}
        \toprule
        \multicolumn{1}{c|}{\multirow{2}{*}{\textbf{Features}}} & \multicolumn{4}{c|}{\textbf{Sample}} & \multicolumn{4}{c}{\textbf{Speaker}}\\
         & \textbf{Accuracy} & \textbf{AUC} & \textbf{TPR$_{\text{FPR} = 0.1}$} & \textbf{TPR$_{\text{FPR} = 0.01}$} & \textbf{Accuracy} & \textbf{AUC} & \textbf{TPR$_{\text{FPR} = 0.1}$} & \textbf{TPR$_{\text{FPR} = 0.01}$}\\ \midrule
        Errors & $ 69.8 \pm 0.4 $ & $ 76.0 \pm 0.2 $ & $ 30.6 \pm 1.1 $ & $ 4.6 \pm 1.2 $ & 
        $ 77.7 \pm 0.1 $ & $ 83.0 \pm 0.3 $ & $ 56.1 \pm 1.0 $ & $ 11.8 \pm 3.0 $\\   
        Losses & $ 86.8 \pm 0.2 $ & $ 93.3 \pm 0.1 $ & $ 78.9 \pm 0.8 $ & $ \mathbf{24.2 \pm 4.0} $ &
        $ 75.9 \pm 0.4 $ & $ 82.1 \pm 0.2 $ & $ 53.8 \pm 2.3 $ & $ 9.2 \pm 2.9 $ \\ \midrule
        Losses + GF & $ 87.3 \pm 0.3 $ & $ 92.8 \pm 0.1 $ & $ 73.8 \pm 0.9 $ & $ 15.2 \pm 2.5 $ &
        $ \mathbf{79.8 \pm 0.3} $ & $ \mathbf{84.8 \pm 0.2} $ & $ \mathbf{63.4 \pm 1.4} $ & $ 13.4 \pm 3.5 $ \\ 
        Losses + AF & $\mathbf{ 88.3 \pm 0.3} $ & $ \mathbf{94.2 \pm 0.1} $ & $ \mathbf{81.6 \pm 1.0} $ & $ 22.5 \pm 4.1 $ & 
        $ 74.9 \pm 0.7 $ & $ 80.6 \pm 0.2 $ & $ 50.1 \pm 1.9 $ & $ 14.6 \pm 2.5 $ \\
        Losses + GF + AF & $ 88.1 \pm 0.2 $ & $ 93.9 \pm 0.1 $ & $ 79.2 \pm 1.5 $ & $ 17.9 \pm 2.8 $ &
         $ 78.1 \pm 0.4 $ & $ 83.5 \pm 0.2 $ & $ 62.6 \pm 1.1 $ & $ \mathbf{14.9 \pm 2.9} $ \\
        \bottomrule
    
    \end{tabular}
     \vspace{-2mm}
    \caption{Results for MI performance for equal target and shadow models (model \textbf{T1}), per feature set at both sample and speaker-level.}
    \label{tab:transLS360_sam_features}
\end{table*}
\begin{table*}[ht]
\setlength\tabcolsep{3 pt}
    \centering
    \begin{tabular}{cl|rrrr|rrrr}
        \toprule
        \multirow{2}{*}{\shortstack[c]{\textbf{Shadow} \\ \textbf{Model}}} & \multicolumn{1}{c|}{\multirow{2}{*}{\textbf{Features}}} & \multicolumn{4}{c|}{\textbf{Sample}} & \multicolumn{4}{c}{\textbf{Speaker}}\\
         & & \textbf{Accuracy} & \textbf{AUC} & \textbf{TPR$_{\text{FPR} = 0.1}$} & \textbf{TPR$_{\text{FPR} = 0.01}$} & \textbf{Accuracy} & \textbf{AUC} & \textbf{TPR$_{\text{FPR} = 0.1}$} & \textbf{TPR$_{\text{FPR} = 0.01}$}\\ \midrule
        \textbf{T2} & Errors & $ 70.1 \pm 0.2 $ & $ 77.3 \pm 0.2 $ & $ 33.2 \pm 1.7 $ & $ 5.6 \pm 1.5 $ & 
        $ 77.1 \pm 0.4 $ & $ 82.8 \pm 0.3 $ & $ 52.3 \pm 1.5 $ & $ \mathbf{8.5 \pm 2.4} $ \\ 
        \textbf{T2} & Losses & $ 86.0 \pm 0.2 $ & $ 92.3 \pm 0.2 $ & $ 75.2 \pm 1.4 $ & $ 0.0 \pm 0.0 $ &
        $ 76.1 \pm 0.4 $ & $ 81.5 \pm 0.2 $ & $ 49.7 \pm 1.7 $ & $ 6.2 \pm 2.7 $ \\
        \textbf{T2} & Loss. + GF + AF & $ \mathbf{86.6 \pm 0.1} $ & $ \mathbf{93.0 \pm 0.1} $ & $ \mathbf{75.7 \pm 0.7} $ & $ \mathbf{12.8 \pm 6.6} $ &
        $ \mathbf{79.7 \pm 0.4} $ & $ \mathbf{84.2 \pm 0.3} $ & $ \mathbf{59.7 \pm 1.8} $ & $ 7.9 \pm 2.3 $ \\ \midrule
        \textbf{C1} & Errors & $61.3 \pm 1.6$ & $67.7 \pm 1.9$ & $22.1 \pm 1.4$ & $2.6 \pm 1.0$ & 
       $ 64.3 \pm 4.7 $ & $ 74.8 \pm 2.6 $ & $ \mathbf{40.0 \pm 4.6} $ & $ \mathbf{6.6 \pm 2.3} $ \\
        \textbf{C1} & Losses & $57.0 \pm 0.6$ & $80.4 \pm 1.3$ & $24.2 \pm 7.9$ & $0.0 \pm 0.0$ &
        $ \mathbf{74.9 \pm 0.7} $ & $ \mathbf{78.6 \pm 0.7} $ & $ 36.2 \pm 2.0  $ & $ 1.6 \pm 0.7 $ \\ 
        \textbf{C1} & Loss. + GF + AF & $\mathbf{69.8 \pm 1.7}$ & $\mathbf{81.3 \pm 1.9}$ & $\mathbf{36.8 \pm 1.3}$ & $\mathbf{6.0 \pm 1.9}$ &
        $ 61.8 \pm 3.6 $ & $ 65.0 \pm 5.8 $ & $ 32.6 \pm 7.3 $ & $ 5.7 \pm 2.2 $ \\
        \bottomrule
    \end{tabular}
    \vspace{-2mm}
    \caption{Results for MI performance for each shadow model (\textbf{T2}, \textbf{C1}), for target model \textbf{T1}, per feature set at both sample and speaker-level.}
    \label{tab:aggregated_results_sam}
\end{table*}
\vspace{-0.2cm}
\subsection{Ablation study}
\vspace{-0.2cm}
The results for our ablation study are aggregated in Table~\ref{tab:transLS360_sam_features}.
The baseline feature set utilising the error features corresponds to a \textit{black-box} scenario, wherein an auditor cannot access model weights or unprocessed outputs (as used in~\cite{shah2021evaluating}).
The results are shown in line 1, for both sample-level (left) and speaker-level MI (right).  

The results obtained for sample-level MI with error features have the lowest accuracy among the considered feature sets, at approximately 70\%.
As lines 2--3 demonstrate, allowing access to the output logits of the model and incorporating loss information enhances all success metrics compared to the black-box scenario.
For sample-level MI, by simply using the losses, results improve by over 15\% for both Acc and AUC when compared to the error features, showing that the loss features can 
separate seen and unseen samples very~well.

When combining the loss features with each of the perturbations, lines 3--4, and their combination, line 5, we observe that all feature sets bring a similar level of improvement, reaching values close to 88\% and 94\% for Acc and AUC.
While the Gaussian features (\textbf{GF}) can be computed at little expense with \emph{grey-box} access to the model, the features based on adversarial samples (\textbf{AF}) require \textit{white-box} access to perform backpropagation.
Though the AF slightly outperform the GF, the GF still provide a good performance. 
This can be advantageous when computational resources are limited, and generating adversarial perturbations is not feasible. 


When considering speaker-level MI, the results are quite different. 
In this case, the loss features alone underperform when compared to the error-based features that achieve an accuracy of 77\%.
The perturbation-based methods are only able to improve upon these results by a margin of $\sim$2\%. 
A possible reason for this contrast is 
that while ASR models are trained to minimise the loss of specific samples, the model's training process does not specifically account for speakers. 
Consequently, loss and error features likely carry similar information regarding the membership of specific speakers.

One might question why the error-based features provide much better results for speaker-level MI than for sample-level MI.
We hypothesise that this is due to how we set up our MI dataset. 
At the sample level, all utterances belong to speakers that are present in the model's training set, making them more challenging to distinguish. 
In this sense, the results we obtain in this work for these features are similar to those obtained by~\cite{shah2021evaluating}. 
This is in contrast to other works, where this distinction is not made, and where negative samples always correspond to unseen speakers, thus simplifying the task and achieving better performances~\cite{li2023membership}.
\vspace{-0.2cm}
\subsection{Shadow model performance}
\vspace{-0.1cm}
Table~\ref{tab:aggregated_results_sam} provides results for the more realistic scenarios where the shadow models are not based on the same dataset as the target model (models \textbf{T2} and \textbf{C1}). 
Here, for brevity, we only provide the results for three feature sets.
For the experiments performed with the transformer shadow model (\textbf{T2}), the results follow a similar trend to the above, with the combination of the loss- and perturbation-based features providing the best overall sample-level results, achieving an accuracy close to 87\%.

However, when the shadow model is based on a different architecture (model \textbf{C1}, a conformer model instead of a transformer model), the accuracy for sample-level MI deteriorates to roughly 70\%. 
Nevertheless, in terms of the AUC, the new feature sets provide an improvement of nearly 15\% when compared to the error-based features. 
In this case, at the speaker level, the best-performing feature set is the set of loss features, with the combination of losses and perturbations achieving the worst results for most metrics. 
A possible explanation for this may be that, given the large differences between architectures, the behaviour of the decision boundary around unseen data points may not be comparable between models, making these perturbations ill-adjusted to this case.

\vspace{-0.2cm}
\subsection{Performance at low FPR operation points}
\vspace{-0.1cm}
In Table~\ref{tab:transLS360_sam_features}, for a maximum FPR of 10\%, the proposed loss- and perturbation-based features reach values above 75\% TPR, more than 40\% above error features. 
Similarly, for the very low value of 1\% FPR, the proposed features outperform error-based features, although with much lower absolute TPRs. 
A similar behaviour is observed for shadow model \textbf{T2} in Table~\ref{tab:aggregated_results_sam}.
This shows that low-FPR operation points, which are particularly relevant to MI auditing \cite{carlini2021membership}, also benefit from loss- and perturbation-based features. 

\section{Conclusions}
\label{sec:conclusions}

In this work, we have explored the advantage of using loss-based features, together with Gaussian and adversarial perturbations, to perform membership inference in ASR models. 
This work was framed under an auditing setting, with the goal of determining if users' speech data was used during model training without their consent. 
With this question in mind, we considered various levels of: access to model outputs; knowledge of the distribution of the target model's training data; and knowledge of the model's architecture.

We performed sample-level and speaker-level experiments. 
At the sample level, the proposed features greatly outperform previously proposed error features.
This occurs even for very low false positive rates, that take into account the importance of wrong decisions in model auditing. 
At the speaker level, the proposed features obtain similar or improved results when compared to the original error features, depending on the feature configuration.
Overall, our results show that easy and computationally cheap features improve MI performance in ASR, particularly for auditing scenarios.

There are several possible avenues for future research.
For instance, it remains to be understood if loss-based features can be applied to shadow models trained with different loss functions. 
Similarly, shadow and target models with very different architectures may have very different loss distributions, making it important to explore techniques that minimise this mismatch.
Exploring the impact of the differences in recording conditions and speaking styles between the shadow/target model's training datasets would also be interesting, as well as exploring
methods to improve TPR scores obtained at very low FPRs, particularly if MI should be used for model auditing. 
We should explore how well our methodology performs when tested on different sub-groups of the population.
In addition, while differential privacy might serve as a theoretical protection against membership inference \cite{hu2022membership}, it also limits the possibility of auditing a model's training data. 
As such, it would be worth exploring the trade-off between what can be considered opposing goals.

\bibliographystyle{IEEEbib}
\bibliography{bibliography}


\end{document}